\documentclass[sigconf]{acmart}
\usepackage{subcaption}
\usepackage{xcolor}
\usepackage{float}

\usepackage{wrapfig}

\definecolor{causal_color}{HTML}{E69F00}
\definecolor{arguably_causal_color}{HTML}{9370DB}

\AtBeginDocument{%
  }

\acmDOI{}
\acmConference[UKAIRS '25]{Make sure to enter the correct
  conference title from your rights confirmation email}{September 08--09,
  2025}{Newcastle upon Tyne, UK}
\acmBooktitle{UKAIRS '25: UK AI Research Symposium,
  September 08--09, 2025, Newcastle upon Tyne, UK}
\acmISBN{}

\begin{document}

% \title{When do causal predictors (not) generalize better to new domains?}
\title{A Shift in Perspective on Causality in Domain Generalization}

\author{Damian Machlanski,$^{1, 2}$
Stephanie Riley,$^{1, 3}$
Edward Moroshko,$^2$
Kurt Butler,$^{1, 2}$
\\ Panagiotis Dimitrakopoulos,$^{1, 2}$
Thomas Melistas,$^{2, 4, 7}$
Akchunya Chanchal,$^{1, 5}$
Steven McDonagh,$^2$
\\ Ricardo Silva,$^{1,6}$
Sotirios A. Tsaftaris$^{1, 2, 4}$}
\email{chai@ed.ac.uk}
\affiliation{%
\small{  \institution{\begin{tabular}{lll}
1. CHAI Hub, UK & 2. The University of Edinburgh, UK & 3. University of Manchester, UK \\ 4. Archimedes, Athena Research Center, Greece & 5. King's College London, UK & 6. University College London, UK \\ & 7. National and Kapodistrian University of Athens, Greece
\end{tabular}}}
  % \city{Ed}
  \country{}
}

\renewcommand{\shortauthors}{CHAI Hub}

\begin{abstract}
The promise that causal modelling can lead to robust AI generalization has been challenged in recent work on domain generalization (DG) benchmarks. We revisit the claims of the causality and DG literature, reconciling apparent contradictions and advocating for a more nuanced theory of the role of causality in generalization. We also provide an interactive demo at \href{https://chai-uk.github.io/ukairs25-causal-predictors/}{this URL}.
\end{abstract}

\maketitle

\section{Introduction}
Generalization is one of the major goals of AI research. In the problem of \textit{domain generalization} (DG), we aim to build models that can leverage information in one environment to make predictions in a different environment. An implicit and necessary assumption for generalization is that the relationship between model inputs and the prediction target is consistent, or \textit{stable}, across environments. Much research on the DG problem has focused on the design of models which are stable predictors, in the sense that they are trained on one or more environments (domains), and their performance is evaluated in a separate set of environments. In other words, the \textit{training} data that the model is optimized on statistically differs from \textit{test} data. The goal for the model is then to perform well on the test data despite those differences. 

The claim that causality relates to the DG problem arose around ideas of invariant prediction \cite[pp.141-142]{peters2017elements} and independent causal mechanisms \cite[pp.16-21]{peters2017elements}. By modelling causal systems as collections of modular, independent mechanisms, it becomes possible to discuss how the statistics of a real-world system would change under interventions. 
Then as long as the causal mechanism in the real world is fixed, the distribution of the target variable conditioned on its parents should be invariant under changes to the rest of the system.  Hence, if we build a predictor that can model the causal mechanism directly, using the causal parents as model inputs to predict the target, then the thought is that this predictor should be robust to a large class of changes, such as changes in environment. 

\section{What could go wrong?}
As suggested by the authors in  \cite{nastl2024causal}, a na{\"i}ve interpretation of causality theory is that to build a stable predictor, the set of features input to the prediction model should be constrained to a certain set of \textit{causal features} determined via causal discovery or from eliciting a human expert. However, empirical evidence from DG benchmarks shows that the performance of such models is consistently overshadowed by models which use all available features \cite{nastl2024causal}. At a glance, this appears to contradict causal theory, but closer inspection of the benchmarks in question tells another story. 

\begin{figure}[t]
  \centering

  \begin{subfigure}[b]{0.49\linewidth}
    \includegraphics[width=\linewidth]{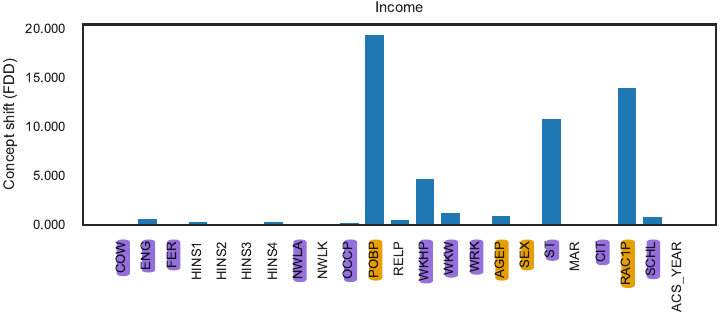}
    \caption{Income dataset}
    \label{fig:income}
  \end{subfigure}
  \hfill
  \begin{subfigure}[b]{0.49\linewidth}
    \includegraphics[width=\linewidth]{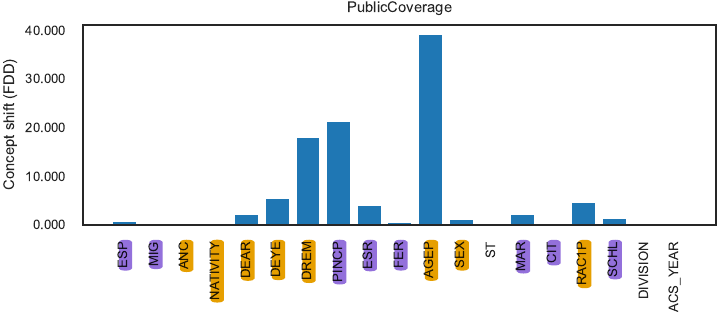}
    \caption{PublicCoverage dataset}
    \label{fig:public_coverage}
  \end{subfigure}

  % \begin{subfigure}[b]{0.49\linewidth}
  %   \includegraphics[width=\linewidth]{figures/poverty.pdf}
  %   \caption{Poverty dataset}
  %   \label{fig:poverty}
  % \end{subfigure}

  \begin{subfigure}[b]{0.49\linewidth}    
    \includegraphics[width=\linewidth]{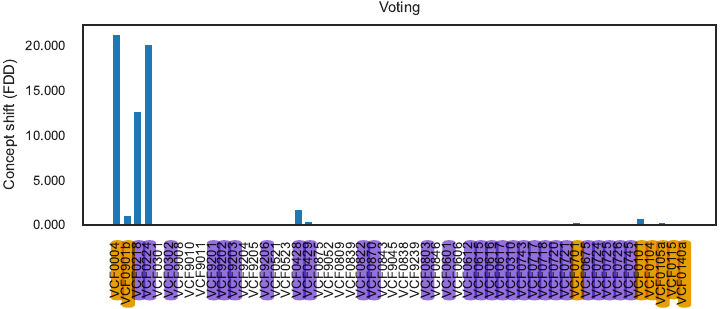}    
    \vspace{0.1em}
    \caption{Voting dataset}
    \label{fig:voting}
  \end{subfigure}
  \hfill
  \begin{subfigure}[b]{0.45\linewidth}
    \includegraphics[width=\linewidth]{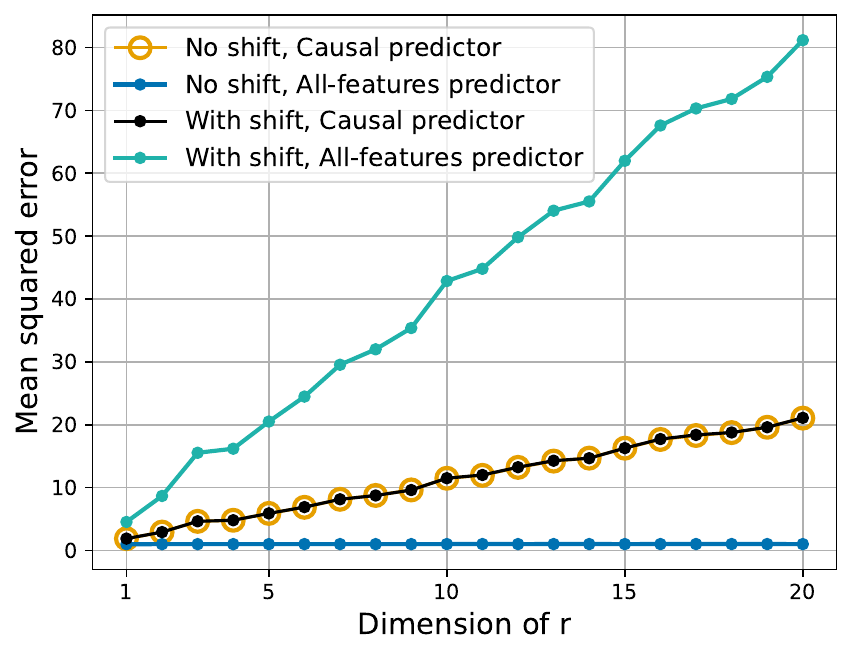}
    \caption{Synthetic example}
    \label{fig:causal_vs_all}
  \end{subfigure}

  \caption{ (a-c) Concept shift for different datasets \cite{nastl2024causal}. \textcolor{causal_color}{Orange} - causal features, \textcolor{arguably_causal_color}{Purple} - arguably causal features. (d) Comparison of causal predictor and all-features predictor on data with and without concept shift, as described in \eqref{eq:synthetic}.}
  \label{fig:concept_shift_all}
\end{figure}

In Figure \ref{fig:concept_shift_all} we present several concept shift plots from appendix E.12 in  \cite{nastl2024causal}, where we have highlighted features on the x-axis according to the authors' classification - orange for causal features, purple for arguably causal features and black (non-colored) for other features.
We selected three datasets where the authors reported the largest performance gaps between causal predictors and all-features predictors (Figure 1 (left) in \cite{nastl2024causal}).

\textbf{Misalignment Between Concept Shift and Causal Features.}
A close examination of Figure \ref{fig:concept_shift_all} reveals a pattern where many features classified as "causal" or "arguably causal" by the authors exhibit the largest concept shifts. This observation contradicts the fact that causal mechanisms should remain invariant across domains. For example, in the Income dataset, features like "POBP" and "RAC1P" show a large concept shift yet are classified as causal. Similarly, in the PublicCoverage dataset, features like "AGEP", "DREM" and "DEYE" that were deemed causal, or "PINCP" that was deemed arguably causal, display significant concept shifts. The fact that their designated causal features show such a large concept shift across domains indicates either incorrect causal attribution or the presence of unmeasured confounders.
% \paragraph{The Advantage of Non-Causal Features}
Observing the pattern of non-causal (black) features across datasets in Figure \ref{fig:concept_shift_all}, many non-causal features exhibit minimal (or zero) concept shift. This stability across domains essentially provides the models with reliable predictive signals that remain stable between training and testing domains. In this context, it is unsurprising that an all-features model outperforms a causal features model, since all-features models benefit from stable non-causal features.

\textbf{Methodological Concerns in Causal Feature Selection.}
The fact that several causal features exhibit concept shift patterns warrants some scrutiny. Because the experiments in \cite{nastl2024causal} did not evaluate concept stability before inclusion into the feature set, it could be argued that there are methodological weaknesses in the study. Rather than testing if "causal predictors generalize better", the results suggest that the assumptions of causal generalization do not hold.

\textbf{A Synthetic Experiment Demonstrating the Impact of Concept Shift.}
% \begin{figure}[H]
%   \centering
%   \includegraphics[width=0.49\linewidth]{figures/causal_vs_all.pdf}
%   \caption{Comparison of causal predictor and all-features predictor on data with and without concept shift.}
%   \label{fig:causal_vs_all}
% \end{figure}
We run a simple synthetic experiment that demonstrates when causal features provide superior out-of-domain generalization. We generated data as follows: 
\begin{equation}
    y=ax+s\textbf{b}^{\top}\textbf{r}+\mathcal{N}(0,1)
    \label{eq:synthetic}
\end{equation}for some fixed $a$ and $\mathbf{b}$, where $x$ is the causal feature, $\mathbf{r}$ is a vector of non-causal features, and $s$ controls concept shift. In the training data we set $s=1$, while in the test data, we examined two cases: $s=1$ (no concept shift) and $s=-1$ (with concept shift). We then trained linear regression models using either only the causal feature $x$, or all features ($x$ and $\mathbf{r}$).
The results appear in Figure \ref{fig:causal_vs_all}. As expected, for $s=1$ (in test data), the all-features predictor outperforms the causal predictor. However, for $s=-1$, i.e. when concept shift is present in non-causal features, the causal predictor outperforms the all-features predictor. When non-causal features experience a concept shift across domains, causal predictors lead to superior out-of-domain generalization. However, in the datasets examined by \cite{nastl2024causal} concept shift in non-causal features is minimal or nonexistent, which explains the authors' findings on why causal predictors performed worse than all-features predictors.

\section{Important considerations}
The previous section showed that achieving good DG performance is not just a matter of feature selection. However, causality can still weigh in on several aspects of the DG problem.

\textbf{Latent confounding and causal mechanisms.}
We emphasize that the causal feature sets in \cite{nastl2024causal} are defined as sets of observed features, not as the set of causes of the target. 
In general, there may be unobserved confounder variables which influence the target, and since they are unobserved they cannot be used in a regression scheme. This is important, since the theory of causal inference shows that causal \textit{mechanisms} are expected to be stable across environments \cite[pp.167-168]{peters2017elements}. Modelling of a causal mechanism requires all of its inputs to be observed, so a relationship confounded by unobserved variables cannot be guaranteed to represent the causal mechanism. 
%Causal theory guarantees generalization of the predictor when all causes are observed, but these claims are not applicable in the presence of confounders. 
In practice, it is rare that one can claim that all causes of the target have been included in the feature set, so environment-varying latent variables are likely to skew predictions. %Models which employ all available features may also learn implicit representations of latent confounders, improving generalization.

% The need to identify invariant (causal) predictors may be motivated by different problems, leading to the development of different methods. Naively applying them to a variety of tasks without careful consideration is rather irresponsible at best.

\textbf{Non-causal relationships are not always unstable.}
From a DG perspective, it is important to note that concepts like stability are context-dependent.
In the cases when input features have the anti-causal relationship with the prediction target (e.g. predicting an illness from symptoms), focusing exclusively on causal predictors discards important information. A spurious (non-causal) relationship is likely not generalizable to all possible environments, but it is possible that it is stable across the finite set of environments considered in a study. Since larger sets of features have a better chance of containing stable features, all-feature predictors may sometimes outperform causal feature predictors in generic settings. 

\textbf{The use of causal discovery in prediction.}
The goal of causal discovery is to establish causal relationships among all dataset variables. This task is often agnostic of any variable categorization, such as predictors or targets. For this reason, the use of such methods is usually not the best choice if the task is to specifically select causal predictors of the target variable.

% \paragraph{Anti-causal graph structures}
% In the cases when input features have the anti-causal relationship with the prediction target (e.g. predicting an illness from symptoms), focusing exclusively on causal predictors is no longer meaningful and rather harmful as it discards important information.

% \paragraph{The presence of latent variables}
% The presence of hidden latent variables can obscure causal relationships among observed variables and lead to erroneous feature selection.

% \paragraph{Availability of causal features}
% An extreme case of this problem occurs when $\mathcal{C}= \emptyset$. In this case, the causal predictor always outputs a constant value. Thus, any features in $\mathcal{F}$ that are correlated with the target can be used to improve prediction. The causal model can only then perform well under domain shift if all features experience a concept shift, which may be unlikely in several cases.

\textbf{Signal-noise ratio.}
Real world data collection is noisy, so features obtained from real data are likely to record noise. If causal features are recorded with a low signal-noise ratio (SNR), then they may not be useful for building a predictor. Similarly, if a signal from the target is strong in spurious variables, then those features may remain useful in prediction. Changes in SNR can also constitute domain shifts that are useful for discovering causal features \cite{salaudeen2024domain}.

\textbf{Strength of the shift.}
If a predictor using spurious correlations exceeds a causal predictor in performance, then a small shift in the domain might not be enough to ruin the advantage of the spurious predictor. As a result, a larger shift in domain is required for the benefits of causal modelling. The spurious correlation reversal condition discussed in \cite{salaudeen2024domain} exemplifies this phenomenon.

\section{Conclusion}
The insights of causality into the DG problem  cannot be reduced to just principles for feature selection. Deeper insights into the roles of confounding, the nature of anticipated shifts, and the availability of stable, non-causal predictors are subjects of future investigation.

\begin{acks}
{We acknowledge support of the UKRI AI programme, and the Engineering and Physical Sciences Research Council, for CHAI - Causality in Healthcare AI Hub [grant number EP/Y028856/1].}
\end{acks}

\bibliographystyle{ACM-Reference-Format}
\bibliography{sample-base}

%\appendix

\end{document}